\documentclass[runningheads]{llncs}
\usepackage{graphicx} 

\usepackage{gensymb}
\usepackage{textcomp}
\usepackage{multirow}
\usepackage{pifont}

\usepackage{tikz}
\usepackage{comment}
\usepackage{amsmath,amssymb} 
\usepackage{color}

\begin{document}
\pagestyle{headings}
\mainmatter
\def\ECCVSubNumber{25}  

\title{
Feed-Forward On-Edge Fine-Tuning Using Static Synthetic Gradient Modules
} 

\titlerunning{Feed-Forward On-Edge Fine-Tuning using SGMs}
%

\author{
Robby Neven\inst{1} \and 
Marian Verhelst\inst{2} \and 
Tinne Tuytelaars\inst{1} \and 
Toon Goedem\'e\inst{1}
}

\authorrunning{R. Neven et al.}
%

\institute{
KU Leuven/ESAT-PSI \and 
KU Leuven/ESAT-MICAS \\ 
\email{first.last@kuleuven.be}
}

\maketitle

\begin{abstract}
Training deep learning models on embedded devices is typically avoided since this requires more memory, computation and power over inference. In this work, we focus on lowering the amount of memory needed for storing all activations, which are required during the backward pass to compute the gradients. Instead, during the forward pass, static Synthetic Gradient Modules (SGMs) predict gradients for each layer. This allows training the model in a feed-forward manner without having to store all activations. We tested our method on a robot grasping scenario where a robot needs to learn to grasp new objects given only a single demonstration. By first training the SGMs in a meta-learning manner on a set of common objects, during fine-tuning, the SGMs provided the model with accurate gradients to successfully learn to grasp new objects. We have shown that our method has comparable results to using standard backpropagation.

\end{abstract}
\section{Introduction}
Most of the embedded devices currently running deep learning algorithms are used for inference only. The model gets trained in the cloud and optimized for deployment on the embedded device. Once deployed, the model is static and the device is only used for inference. If we want to incorporate new knowledge into the device, the standard way is to retrain the model in the cloud.

However, having an agile model that can be retrained on the embedded device has several advantages. Since the device is the origin of the data, no traffic between the device or the cloud has to be established, reducing latency and bandwidth. One example of this are self-driving cars, which can fine-tune their model to changing environments on the fly without the need to interact with a server or the cloud.  Also, some applications like surveillance cameras deal with privacy issues which make it impossible to centralise the data. Therefore, the data has to be kept local, forcing the model to be trained on the device. One other advantage is that each device can fine-tune its model to its own specific input data, resulting in personalized models. Instead of training one large generic model, each device can train a smaller specialist model on their specific input data, which can be beneficial (e.g., cameras only seeing one specific viewpoint).

Nevertheless,  training on embedded devices is avoided for several reasons. The most determining one is that most of the embedded devices are severely resource constrained. A standard training loop requires substantially more memory, computations, precision and power than inference, which are not available on the embedded device. Another constraint is the lack of supervision, which enforces the use of unsupervised or semi-supervised methods. Having to deal with all these constraints makes training on the embedded device nearly impossible, or at the least very cumbersome.

The goal of this paper is to investigate the training cycle on embedded systems, while focusing on the memory constraint. Indeed, in small embedded platforms like \cite{Verhelst2020}, the amount of memory appears to be the bottleneck. More specifically, our aim is to lower the total amount of memory by training the network without backpropagation, but in a feed-forward manner. In this way, we address the update-locking problem, referring to the fact that each layer is locked for update until the network has done both a forward and a backward pass. During the forward pass all activations are computed, which are used during the backward pass to calculate the gradients. All layers are locked until their gradient is computed. Having to store all activations for the backward pass is a huge memory burden. If we can train the network in a feed-forward manner, the embedded device only has to store activations of the active layer instead of the whole network.

\begin{figure}[tb]
    \centering
    \includegraphics[scale=0.25]{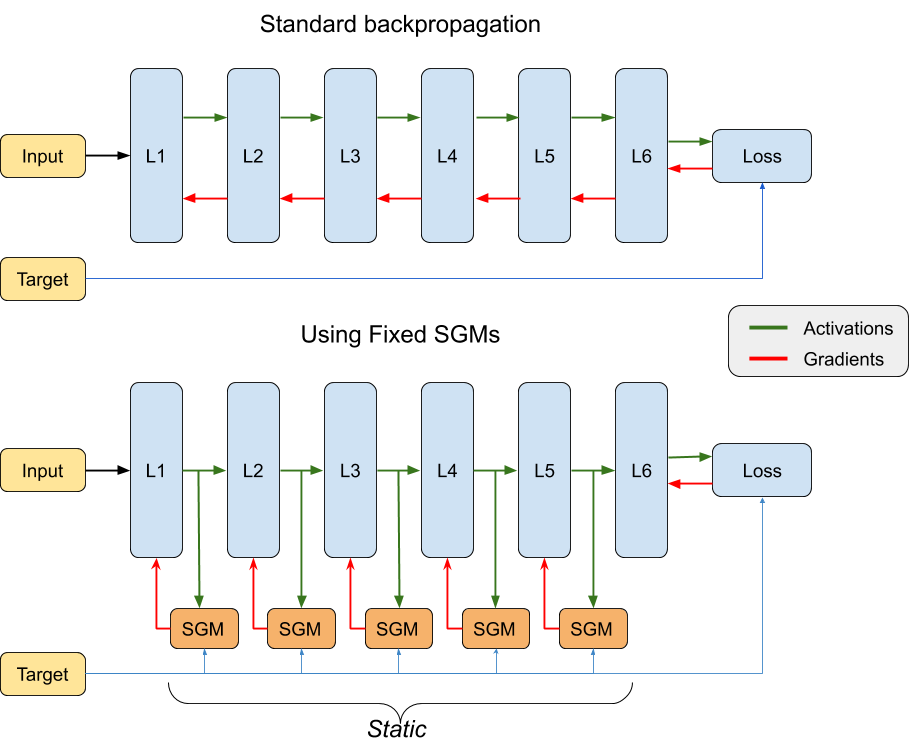}
    \caption{Comparison between standard backpropagation (BP) and using static Synthetic Gradient Modules (SGMs). BP calculates layer activations during the forward pass, which are being used to calculate the gradients during the backward pass. In contrast, when using static SGMs, the gradients can directly be estimated based on each layer's activations and the input label. Therefore, the network can be trained in a feed-forward manner, bypassing the need to store the activations of the whole network.}
    \label{fig:overview}
\end{figure}

Figure~\ref{fig:overview} illustrates our approach. In contrast to standard backpropagation (BP), where the gradient flows backwards through the network, feed-forward training utilizes a gradient directly after computing each layer during the forward pass.  This was already introduced by target propagation algorithms like \cite{frenkel2019learning}, where they estimated this gradient by a fixed random transformation of the target. However, they showed this only worked well in a classification setting, where the target contains the sign of the global error of the network. 

Instead of a fixed random transformation of the label, we propose to estimate a layer’s gradient by transforming the activations of that layer together with the label. This method, called synthetic gradients, was first introduced in \cite{DBLP:journals/corr/JaderbergCOVGK16}. However, their method requires the gradient estimators (which from now on we will refer to as Synthetic Gradient Estimators or SGMs) to be trained in an online manner with a ground truth gradient. Instead, we will use static SGMs, similar to the fixed transformations of \cite{frenkel2019learning}, which are pretrained on a set of tasks and can be used to fine-tune a model to a similar---but unseen---task in a feed-forward manner.

We tested our method on a robot grasping scenario from \cite{gent_robot}, where a robot, equipped with a camera, needs to learn to grasp new objects given only a single image. Since the CV algorithm of the robot runs on an embedded device, this application can greatly benefit from the feed-forward training method. The objective of the model is to estimate the center and rotation of the object, meaning the setup can be seen as a multi-task setting: both a classification and regression problem. Therefore, this setup is ideal to show how our method can cope with different tasks. In the experiments and results section, we will show that we were able to successfully train the model to grasp new objects in a feed-forward manner using the static SGMs. By pretraining the SGMs in a meta-learning manner on a set of common household objects, the static SGMs were able to provide accurate updates to the model to learn new similar objects.  We compared our method to standard backpropagation and showed only a slight accuracy drop.

To summarize, we used static SGMs to train a network in a feed-forward manner. Providing each layer with a gradient during the forward pass, this bypasses the update-locking problem and releases us from the memory burden of storing all activations. The SGMs were pretrained using a meta-learning setup and tested on a robot grasping scenario where we showed similar results to standard backprop.
\section{Related Work}
\subsection{Training With Synthetic Gradients}
The idea of training a network with synthetic gradients generated by small models was introduced by \cite{DBLP:journals/corr/JaderbergCOVGK16}. The goal of their work was to train a distributed network without wasting computational power. When you have a network deployed over multiple devices, the device executing the first layers is going to be idle during the forward pass of the rest of the network. Even when doing the backward pass, the first device has to wait for the backpropagation through all other devices before it gets its downstream gradient. This problem is commonly referred to as the update-locking problem. Their idea was to decouple the forward and backward pass by estimating the gradient with a small neural network. This way the device could update its layers without waiting for the rest of the forward and backward pass. Each layer or part of the network has its own gradient generator, one or two fully connected or convolutional layers, which transforms the activations of that layer into a sufficient gradient. Not surprisingly, they also showed that the quality of the gradient increases when the gradient network also receives the input label.

However, these gradient networks cannot predict gradients without any knowledge. To train these networks, during the forward pass of one batch of inputs, simultaneously, the previous batch gets backpropagated. Each transaction between devices exchange both the forward pass activations as well as the backpropagated gradient. This backpropagated gradient serves as ground truth for the gradient networks which they simply trained with an l2-loss. As shown in \cite{DBLP:journals/corr/JaderbergCOVGK16}, as the target gradient distribution shifts, the gradient network follows this distribution. Therefore, there will always be some lag between the ground truth and predicted gradients, but they showed that this still is able to train the network sufficiently.

\subsection{Training Without Backpropagation}
Besides the use of synthetic gradients, other research explore different substitutes for standard backpropagation, more specifically the biologically plausible methods. One of these methods is the feedback alignment algorithm \cite{lillicrap2014random}. They argue that standard backpropagation in the brain is not possible because the forward and backward paths in the brain are physically separated (weight transport problem). To mimic this behaviour, they used fixed random feedback weights during the backward pass and showed that it could enable the network to learn.

One extension of this work is the direct feedback alignment method (DFA) \cite{nkl2016direct}. Instead of backpropagating the error using fixed random weights, they use the fixed feedback weights to transform the final error of the network into a gradient for each layer. Therefore, once the error of the network is computed, each layer gets it gradient at the same time as a random transformation of the loss. One big advantage of this method is that after the forward pass, all layers can be updated in parallel. This somewhat solves the update-locking problem. However, this would require a large increase in hardware. Nevertheless, recent work showed that these algorithms scale to modern deep learning tasks \cite{Lechner2020Learning}, \cite{launay2020direct}, \cite{Moskovitz2018FeedbackAI}, \cite{Han2019DirectFA} and \cite{han2020extension}. Other work have also focused on reducing the fixed feedback weights with sparse connections \cite{Crafton2019DirectFA} and the effect of direct feedback alignment on shallow networks \cite{Illing2019BiologicallyPD}.

All these previous methods use some transformation of the error of the network to provide a gradient to the network. \cite{frenkel2019learning} showed that we don't even need the final error of the network to train the upstream layers. They showed that, only the sign of the error is enough information to train the label. Having the fact that the target for classification tasks already contains the sign (one-hot encoded), they used the target as a proxy for the error of the network. Meaning, instead of computing the final error, each layer receives a random projection of the target as gradient. They showed that this indeed can support learning. This method has some huge advantages for hardware, since the network can now be trained in a feed-forward manner and reduces memory footprint since this solves the update-locking problem. Similar work have been conducted in \cite{alex2020theoretical} and \cite{Manchev2020TargetPI}.

\subsection{Fine-Tuning Models With Meta-Learning}
If we compare the use of synthetic gradients with the bio plausible algorithms, there is not much difference between them. The first uses a network to generate gradients, which is constantly being trained with ground truth gradients. The second uses a fixed random projection of the final error or even the target. The main advantage of the synthetic gradients is that the network behaves in a feed-forward manner, solving the update-locking problem. However, these networks need to be trained. The only bio plausible algorithm from the ones discussed that really solves the update-locking problem is \cite{frenkel2019learning}, however this works only well in classification problems and has worse performance than the synthetic gradients. For this reason, we want to combine the two by using the synthetic gradient setup, but with fixed parameters. Since this relieves us from training the synthetic gradient generators, the network can completely update during the forward pass. However, it is obvious the initialization of the gradient generators is of vital importance.

From \cite{DBLP:journals/corr/JaderbergCOVGK16}, we saw that the distribution of the gradients and activations shift and that the gradient generators follow this distribution while they are trained with ground truth gradients. It is very difficult to pretrain a network which would incorporate this behaviour. Therefore, we only focus on fine-tuning a model to a new task. During fine-tuning, we expect that the distribution of the gradient does not shift that much, making it easier for the fixed gradient generators to be initialized.

To initialize the SGMs, we draw inspiration from the meta learning setting. More specifically, in \cite{DBLP:journals/corr/FinnAL17} they introduced MAML, a model-agnostic meta learning algorithm which enables a model to be fine-tuned with fewer iterations. During training, the algorithm searches for a set of parameters which are equally distant to the optimal parameters of different tasks. If we want to fine-tune the model to a task which is similar to the ones trained on, the initialization parameters are closer to the optimal setting, resulting in fewer gradient descent steps compared to a completely random initialization. We believe that, we can use this algorithm to not only find an initialization for the model, but also for the SGMs.
\section{Method}
To test our method of fine-tuning a model in a feed-forward manner with static SGMs, we adapt the robot grasping setup from \cite{gent_robot} and \cite{DECONINCK2020103474}. The goal of their setup is to fine-tune the model to learn to grasp new objects. Currently, they fine-tune the model on a GPU server. Therefore, this setup is ideal for our method, since our method can lower the total activation memory needed on the embedded device. In this section, we will first explore the robot grasping setup in more detail. Next, we will discuss the use of static SGMs and how these can be initialised by pretraining them using meta-learning.

\subsection{Robot Grasping Setup}
\begin{figure}[tb]
    \centering
    \includegraphics[scale=0.3]{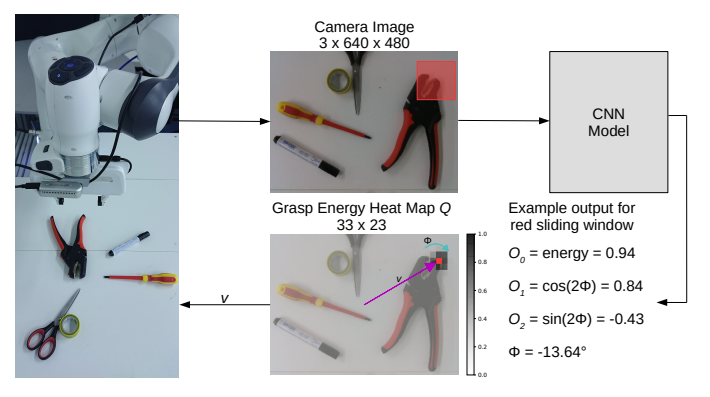}
    \caption{Grasping setup from \cite{gent_robot}. A cobot equipped with a camera runs a CNN model to predict a lower resolution heatmap. The heatmap contains three layers: a grasp quality layer and the rotation angle encoded in a sine and cosine layer. (Figure adapted from \cite{gent_robot})}
    \label{fig:grasping_setup}
\end{figure}

To learn to grasp new objects, a collaborative robot (cobot) equipped with a camera is positioned above the new object. Next, the camera will take one shot of the object, having a frame where the object is centered and in the right angle. Then, to actually grasp the object, a demonstrator guides the end effector of the cobot to the correct grasping position. This uses the "program by demonstration" feature of the cobot, meaning the cobot can execute the same sequence of steps in future grasps. The goal of the computer vision algorithm is to position the cobot right above the object and rotated in the same angle as the demonstration frame. To achieve this, a fully convolutional network (\cite{DBLP:journals/corr/LongSD14}) takes in the camera image and outputs a displacement vector along with a rotation angle, which is provided to a controller as shown in figure \ref{fig:grasping_setup}. Instead of using a standard Cartesian controller, this results in a closed-loop "smart" controller. In this paper we will only focus on the computer vision task.

\subsubsection{The Model}
\begin{figure}[tb]
    \centering
    \includegraphics[width=\textwidth]{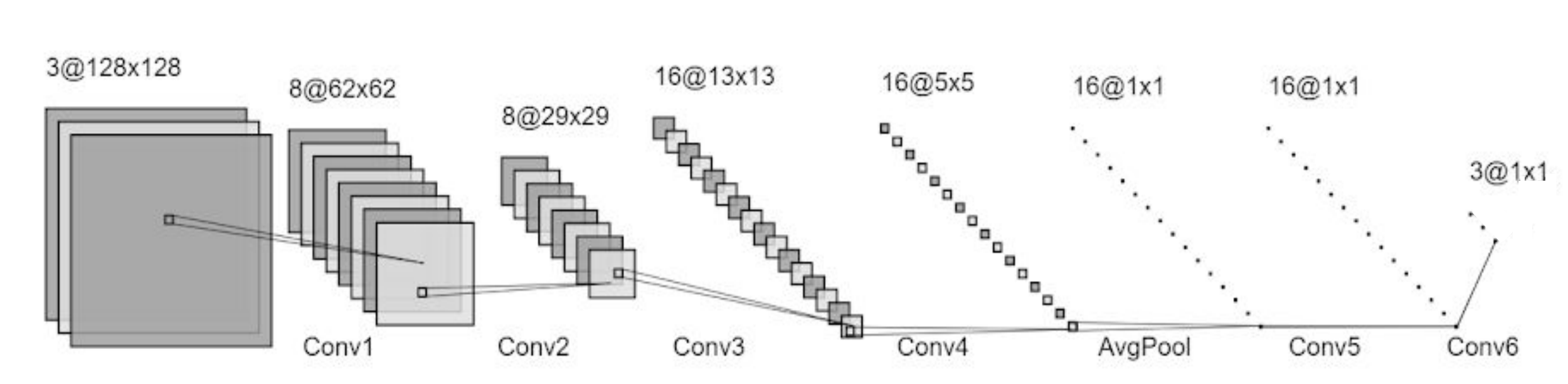}
    \caption{The model used in \cite{gent_robot}. The model works in a fully convolutional manner meaning the last layers are convolutional instead of dense. This way the model can be trained on crops of objects. During deployment, the model operates on full camera images and outputs a lower resolution heatmap as in figure \ref{fig:grasping_setup}}
    \label{fig:graspnet_model}
\end{figure}

During deployment, the model takes in a 3x640x480 image and outputs a heatmap of a lower resolution (3x33x23). Each pixel of the heatmap consists of a grasp quality score and a rotation angle (encoded in sine and cosine). As the network works in a fully convolutional manner (last layers are convolutional instead of dense layers), the model (figure \ref{fig:graspnet_model}) can be trained on crops of size 3x128x128 and output a 3x1x1 pixel. The grasp quality score resembles the centered position of the object in the crop (binary classification), the sine and cosine resembles the rotation angle with respect to the demonstration image (regression).

To learn both the binary classification and angle regression tasks, the grasp quality layer is trained by a log loss while both the rotation layers (sine and cosine) are trained with an l2 loss as in \cite{angles}. Since the end effector of the robot is bipodal, the rotational range of the cobot is $\pm\frac{\pi}{2}$. We incorporate this during the data generation by mapping all angles to this range. Also, to facilitate learning, we limit the output of the angle layers by adding a tanh and sigmoid layer to the sine and cosine respectively, limiting their range to $(\pm 1)$ and (0, 1).

\subsubsection{Dataset}
\begin{figure}[tb]
    \centering
    \includegraphics[scale=0.25]{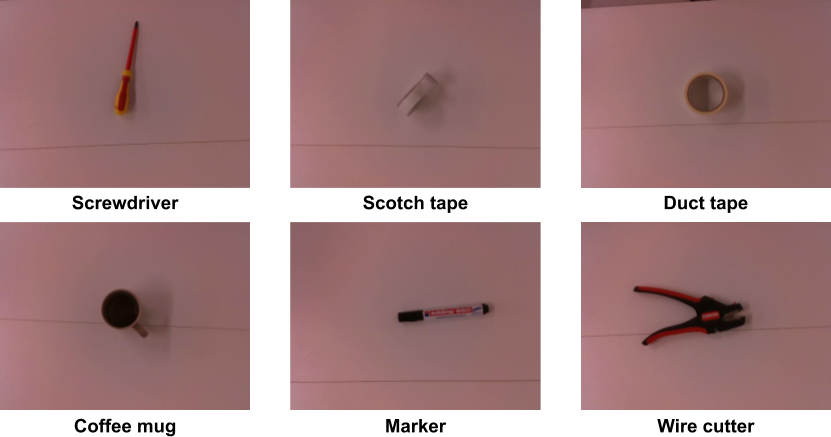}
    \caption{Common household objects from \cite{gent_robot}. Out of the single image of each object, positive and negative samples are generated.}
    \label{fig:dataset}
\end{figure}

In the setup from \cite{gent_robot} they generate positive and negative samples from only one camera frame of a new object. Some frames of the dataset are depicted in figure \ref{fig:dataset}. Positive examples are 128x128 crops, where the center of the object is in the center of the crop. Also, the object is randomly rotated so that the model can learn the rotation of the object during deployment. Negative examples are crops where the object is randomly rotated, but not situated in the center of the crop. Also, all crops use random color jitter and brightness to augment the data for better performance. In contrast to \cite{gent_robot}, instead of generating random samples on the fly, we use a fixed dataset of 1000 images per object: 800 train and 200 validation images.

\subsection{Synthetic Gradient Modules}
To fine-tune the model in a feed-forward manner, we will use SGMs to generate synthetic gradients for each layer as in \cite{DBLP:journals/corr/JaderbergCOVGK16}. However, during fine-tuning, these will remain static. The SGMs will estimate the gradient for each layer based on the activations and the target. It is obvious the SGMs will provide computational overhead. However, for the first layers of the network, the inputs are downscaled drastically so that the activations are smaller than the layer inputs. Since SGMs operate on the activations, the computational overhead can be smaller than computing the layer. In deep networks, it may not be feasible to insert an SGM after each layer. One way to limit the overhead of the SGMs is to insert them after blocks of layers, providing a whole block with a gradient, which is then trained with standard backpropagation. In this case, only one SGM has to be computed, with the downside of storing the activations of the whole block in memory.

Since the network (figure \ref{fig:graspnet_model}) consists of only six layers, each layer will have their own SGM. Therefore, we can use five SGMs (the last layer gets its gradient directly from the loss). Since all the activations are outputs of convolutional layers, our SGMs also consist of convolutional layers, so that they don't take up that much parameters in contrast to fully connected layers. The SGMs are implemented in Pytorch \cite{NEURIPS2019_9015} as a regular layer, except that during the forward pass they compute and store the synthetic gradient. The output of the layer is the input, meaning during the forward pass it acts as a no-op. During the backward pass however, a hook function is executed where the gradient on the input is replaced with the synthetic gradient. This way, the optimizer will use the synthetic gradient to update the layer. During the pretraining step, the hook function will also update the parameters of the SGM, by using the incoming gradient of the output (which is the ground truth gradient for the activations, since the SGM acts as a no-op during the forward pass) as supervision. In all our experiments, the SGMs consist of two convolutional layers with a batchnorm and relu activation layer in between. Our experiments showed that adding the batchnorm layer resulted in a more stable training. Since during fine-tuning the SGMs are static, the batchnorm is only a shift and scale and can be folded into the convolution layer.

\subsection{Training the SGMs}
\begin{figure}[tb]
    \centering
    \includegraphics[scale=0.2]{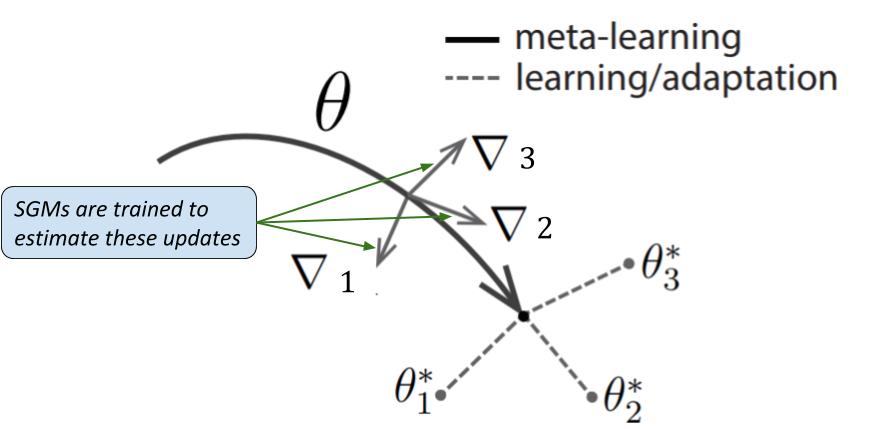}
    \caption{The SGMs are pretrained using MAML. The difference with standard MAML is that we now use the SGMs to update the model instead of standard backpropagation. During training, the SGMs are able to train their parameters with the ground truth gradients like in \cite{DBLP:journals/corr/JaderbergCOVGK16}. This way, the SGMs learn to adapt the model to different tasks. After pretraining, we have shown that these SGMs can be left fixed during deployment to adapt the model to an unseen task, as long as the task has similarities to the training tasks. (Figure adapted from \cite{DBLP:journals/corr/FinnAL17})}
    \label{fig:maml}
\end{figure}

To train the SGMs, we will use the meta-learning algorithm MAML: model-agnostic meta learning \cite{DBLP:journals/corr/FinnAL17}.  The goal of the MAML algorithm is to find an optimal set of model parameters so that the model can be fine-tuned to a new task with as few iterations as possible. This results in finding a set of parameters which are equally distant to the optimal set of parameters for each individual task (see figure \ref{fig:maml}). However, instead of using standard backpropagation to train the model, we will use SGMs.

During the MAML step, the model is trained on different tasks individually, while minimizing the summed error on all tasks. Each task consists of training the model on one object like in figure \ref{fig:dataset}. Since the model is trained by the SGMs, the SGMs learn to provide gradients to fine-tune the model towards all the different objects. Since the goal of MAML is to find parameters which can quickly adapt to new tasks, at the end of the meta-learning step, the model parameters are close to the optimal set of parameters for each task, while the SGMs have learned to provide the gradients to fine-tune the model on all these different tasks.

We will use the end state of the SGMs after the MAML step as initialization. During fine-tuning, these will remain static. Since the goal of MAML is to initialize the model with parameters close to the optimal parameters, during fine-tuning, the gradient distribution will not shift significantly, meaning the SGMs can indeed remain static. However to prove this claim, in the experiment section we will compare the performance of fine-tuning the model using static SGMs and letting them update with a ground truth gradient. 

\section{Experiments and Results}
\subsection{Fine-Tuning Using Standard Backpropagation}
\begin{figure}[tb]
\centering
\includegraphics[scale=0.35]{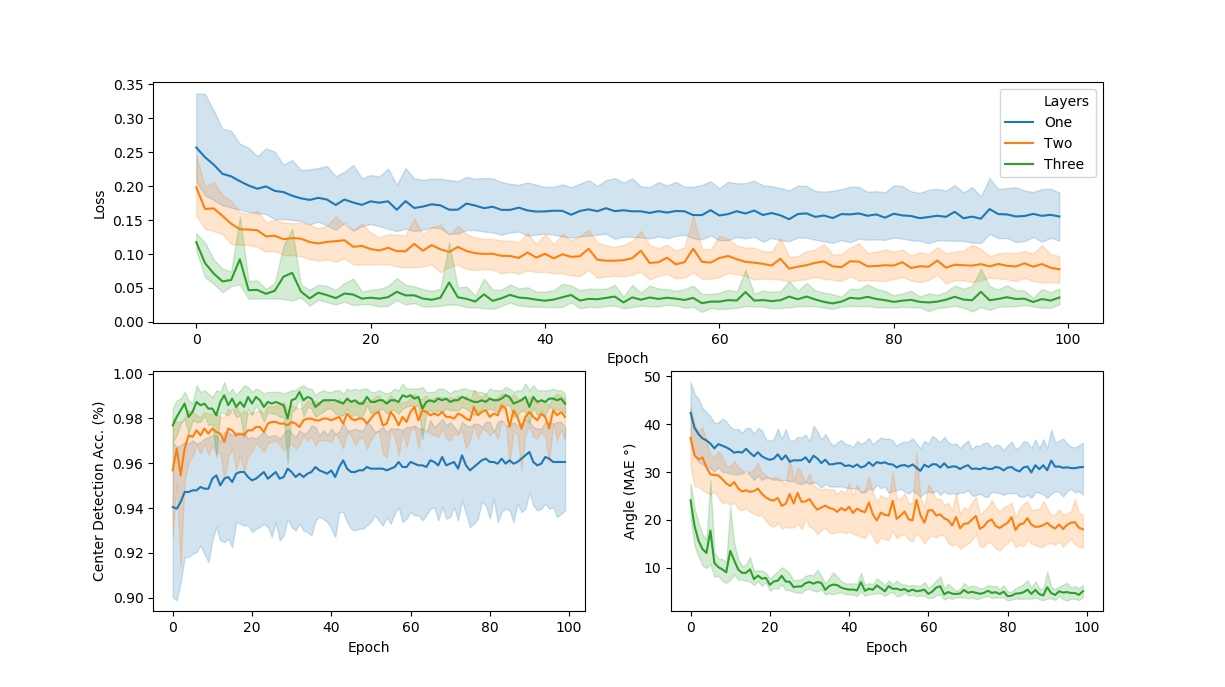}
\caption{Baseline using standard backpropagation. Comparison between fine-tuning the last one, two or three layers of the model. Showing the loss, detection accuracy and mean absolute error averaged for fine-tuning on six different objects' test set. For all objects, fine-tuning three layers has the highest performance. Fine-tuning more than the last three layers resulted in minimal performance increase}
\label{fig:results_backprop}
\end{figure}

As a baseline, we compared the performance of our method to standard backpropagation. For each object in the dataset, we first pretrained the network on the remaining objects. Next, we fine-tuned the model on the object itself. For these experiments we used the Adam optimizer \cite{adam} with a learning rate of $1\times10^{-3}$ with a batch size of 32. We compared the amount of fine-tuned layers ranging from only fine-tuning the last layer to all layers. Figure \ref{fig:results_backprop} shows the result for fine-tuning the last three layers. We noticed that fine-tuning more than the last three layers have little performance increase. Notice also that grasp quality (binary classification task) is already after a couple of epochs close to 100\% accuracy, while the angle regression is a much harder task to learn and reaches a minimum after tens of epochs later.

\subsection{Fine-Tuning Using Static SGMs}

We will insert SGMs after each layer (conv-bn-relu) of the network. The input of these SGMs consists of the activation of the network layer along with the label. Since in our setup the label consists of a binary grasp quality score and an angle encoded in a sine and cosine, we found out that adding three extra channels to the input activations works best: each channel filled with the grasp quality, sine and cosine. Initially, we tried to train the model without using the label as input to the SGMs. While this was able to learn the grasp quality, it was not able to learn to regress the rotation angle.   

\subsubsection{Training the SGMs}
To train the SGMs during the meta-learning step, we experimented with both the l1 and l2 loss functions. While both losses were able to train the SGMs to provide accurate gradients for fine-tuning the model, we did notice an increase in accuracy when using the SGMs trained with the l2-loss. We use standard stochastic gradient descent with a learning rate of $1\times10^{-1}$ and momentum of 0.9 to train both SGMs and the model. 

\subsubsection{Fine-Tuning Using Static SGMs}
In these experiments we initialized the model and SGMs with their pretrained state while keeping the SGMs static. During the forward pass, each layer gets its gradient directly from its designated SGM, fine-tuning the model in a feed-forward manner. Again, both SGD and Adam optimizer were compared with SGD performing slightly better with the same learning rate as during the pretraining step of $1\times10^{-1}$. 

\begin{figure}[tb]
\centering
\includegraphics[scale=0.35]{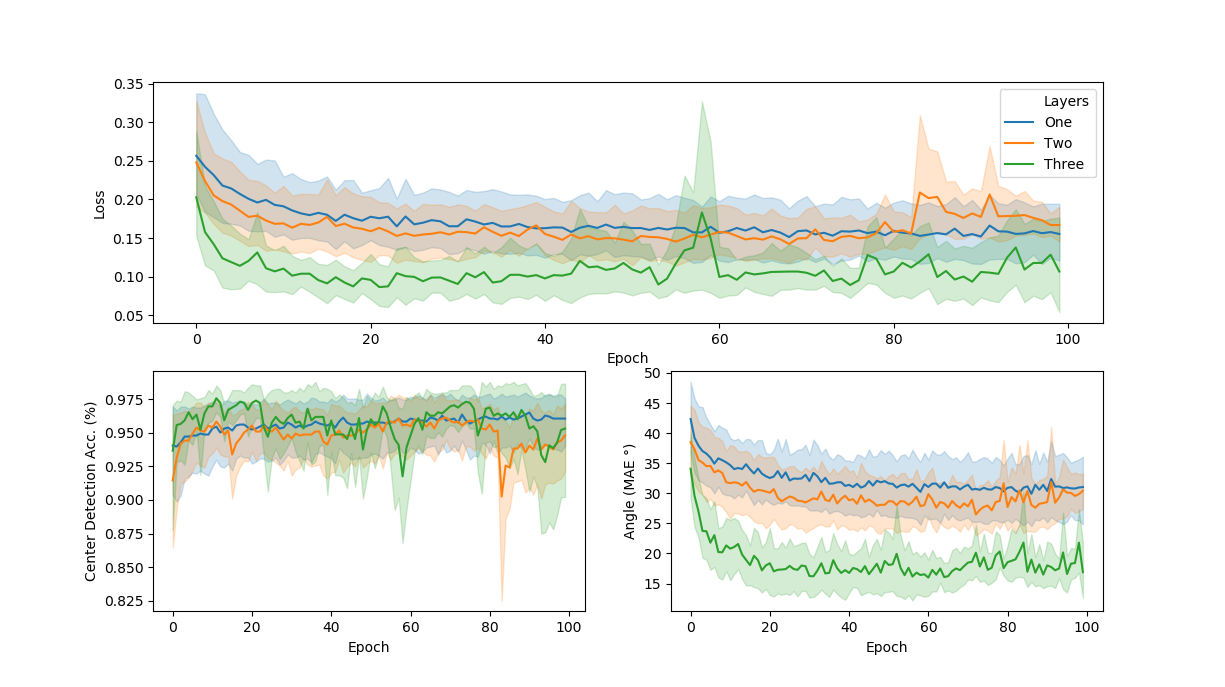}
\caption{Fine-tuning the model using fixed SGMs. Comparison between fine-tuning the last one, two or three layers of the model. Showing mean results for each object's test set. As with standard backprop, fine-tuning the last three layers has the highest performance. However, for all three cases, the detection accuracy is lower and angle error higher than using standard backprop (figure \ref{fig:results_backprop}) by a small amount. While with standard backprop the accuracy did not increase when fine-tuning more than three layers, the performance dropped when using static SGMs (not shown in the graphs).}
\label{fig:results_fixed_sgms}
\end{figure}

\begin{figure}[tb]
    \centering
    \includegraphics[scale=0.3]{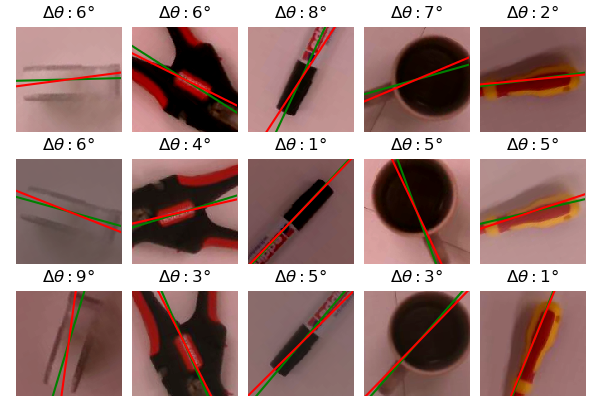}
    \caption{Angle regression of the model fine-tuned using the fixed SGMs on different objects (positive samples from each object's test test). We see that the angle error between the ground truth (green) and predicted (red) angles is between 0-13 degrees. For this setup, the error rate is negligible for a successful grasp.}
    \label{fig:angle_labels}
\end{figure}

Figure \ref{fig:results_fixed_sgms} shows the model is able to be fine-tuned to new objects using the static SGMs. As with standard backpropagaton, the classification task (detection) is able to learn very quickly, while the angle regression needs to train considerably more epochs. The figure also shows that the combined loss converges slower than standard backprop. There also is a significant difference in the MAE of the angle. However, for grasping simple household objects with a bipodal end effector, MAE differences of around 10 degrees will most of the time result in a successful grasp as seen in figure \ref{fig:angle_labels}. Fine-tuning the last three layers resulted in the highest accuracy for both the classification and regression task. When fine-tuning more than three layers, the accuracy dropped below the three-layer experiment. Detailed results can be found in table \ref{tab:fixed_sgms}.

\begin{table}[tb]
\centering
\caption{Comparison between standard backpropagation (BP) and using fixed SGMs when fine-tuning one, two or three layers of the model. Showing both grasp detection accuracy and mean absolute error of the rotation angle on the test set for each object (a-f).}
\label{tab:fixed_sgms}

\begin{tabular}{c|c|c|cccccc|cc}
 & \textbf{Layers} & \textbf{Method} & \textbf{a} & \textbf{b} & \textbf{c} & \textbf{d} & \textbf{e} & \textbf{f} & \textbf{Avg} & \textbf{Std} \\ \hline
\multirow{5}{55pt}{\textbf{Center Detection (\%)}} & 1                  & BP  & 0.92 & 0.98 & 0.98 & 0.98 & 0.99 & 0.98  &  0.97 &  0.02 \\ \cline{2-11} 
                                           & \multirow{2}{*}{2} & BP  & 0.97 & 0.98 & 0.99 & 0.99 & 0.99 & 1.00 &  0.99 &  0.01 \\
                                           &                    & SGM & 0.92 & 0.97 & 0.98 & 0.99 & 0.99 & 0.99 &  0.97 &  0.02 \\ \cline{2-11} 
                                           & \multirow{2}{*}{3} & BP  & 0.99 & 0.99 & 1.00 & 0.99 & 1.00 & 1.00  &  1.0 &  0.0 \\
                                           &                    & SGM & 0.97 & 0.99 & 1.00 & 0.99 & 0.99 & 0.99 &  0.99 &  0.01 \\ \hline \hline
\multirow{5}{55pt}{\textbf{Angle Regression (MAE °)}} & 1                  & BP  & 37.19 & 23.04 & 33.72 & 25.62 & 35.49 & 16.20 &  28.71 &  6.99 \\ \cline{2-11} 
                                           & \multirow{2}{*}{2} & BP  & 18.19 & 17.65 & 13.55 & 20.56 & 20.76 & 6.63  &  16.44 &  4.57 \\
                                           &                    & SGM & 29.21 & 21.34 & 21.09 & 24.37 & 33.25 & 17.48 &  24.07 &  5.01 \\ \cline{2-11} 
                                           & \multirow{2}{*}{3} & BP  & 5.71 & 2.44 & 3.82 & 2.73 & 3.02 & 3.04 &  3.45 &  1.01 \\
                                           &                    & SGM & 16.82 & 8.59 & 15.32 & 14.35 & 9.99 & 13.71 &  12.65 &  2.94 \\ \hline
\end{tabular}
\end{table}

\subsubsection{Trained vs. Random Initialisation}
In the target propagation methods like \cite{frenkel2019learning} and \cite{alex2020theoretical}, the gradient is generated by a fixed transformation of the target. Since the SGMs also have the target as input, one could argue a random initialization might be able to train the network. However, to show the contribution of the trained state, we ran experiments using random static SGMs. We concluded that with a random state, the SGMs were not able to provide the model with accurate gradients, failing to converge on both the classification and regression task.

\subsection{Static vs. Non-Static SGMs}
The main goal of this paper is to fine-tune a network with static SGMs on an edge device, eliminating the need of storing all activations during the forward pass (update-locking problem) . This has the advantage that during deployment the SGMs can provide the gradients on the fly. However, since these are trained to fine-tune the model to different tasks, the gradients must generalize and differ from the ground truth gradients. Therefore, in a last experiment we wanted to show the effect of further training the SGMs during fine-tuning with the ground truth gradient. More specifically, we wanted to show if the trained initialization for the SGMs has any benefit when they are allowed to further update. Table \ref{tab:grasping_stats} summarizes three training methods: (i) with static pretrained SGMs,
(ii) with updated pretrained SGMs,  and (iii) with updated randomly initialized SGMs.

\begin{table}[tb]
\centering
\caption{Comparison between fine-tuning two layers using static or non-static SGMs. Enabling the trained SGMs to further train during fine-tuning results in a slight accuracy improvement. However, when the SGMs are able to further train, there is no clear difference between a pretrained or random initialization.}

\begin{tabular}{c|c|c|cccccc|cc}
                                                   & \textbf{Init.}  & \textbf{Static} & \textbf{a} & \textbf{b} & \textbf{c} & \textbf{d} & \textbf{e} & \textbf{f} & \textbf{Avg} & \textbf{Std} \\ \hline
\multirow{3}{55pt}{\textbf{Center Detection (\%)}}    & \multirow{2}{*}{Trained} & \checkmark       & 0.92       & 0.97       & 0.99       & 0.99       & 0.99       & 0.96       &  0.97   &  0.03     \\
                                                   &                          & \ding{55}       & 0.96       & 0.98       & 0.98       & 0.99       & 0.99       & 0.98       &  0.98   &  0.01     \\ \cline{2-11} 
                                                   & Random                   & \ding{55}        & 0.97       & 0.98       & 0.98       & 0.99       & 0.99       & 0.97       &  0.98   &  0.01     \\ \hline \hline
\multirow{3}{55pt}{\textbf{Angle Regression (MAE °)}} & \multirow{2}{*}{Trained} & \checkmark       & 29.21      & 21.34      & 24.37      & 33.25      & 17.48      & 21.77      &  24.57  &  5.25     \\
                                                   &                          & \ding{55}        & 26.20      & 20.05      & 25.54      & 28.95      & 8.30       & 19.58      &  21.44  &  6.76     \\ \cline{2-11} 
                                                   & Random                   & \ding{55}        & 23.32      & 18.87      & 26.53      & 30.56      & 13.27      & 22.22      &  22.46  &  5.48     \\ \hline
\end{tabular}

\label{tab:grasping_stats}
\end{table}

Table \ref{tab:grasping_stats} shows that enabling the SGMs to further train during fine-tuning, the model achieves higher accuracy. This is obvious, since the SGMs can now follow the ground truth distribution of the gradients. However, the difference between the pretrained and randomly initialized SGMs is remarkable. Having the latter to perform equally well, meaning the pretrained state does not contribute that much. This possibly means that since the SGMs are only a couple of parameters, the SGMs can quickly adapt to the new target gradient distribution, both for the pretrained as well as the random initialization.

\subsubsection{Impact on Memory}
The goal of using static SGMs is to minimize the activation memory to the activation size of the largest layer. When enabling the SGMs to further train during the fine-tuning stage instead of remaining static, we also need to store its activations. Since the activations of the SGMs are the same size as the activations of the layer it provides gradients for, one can optimize which layer can update or freeze its SGM. This is something we did not investigate further in this paper.

\subsection{Memory Advantage}

As we have shown in this paper, having the static SGMs to provide gradients during the forward pass allows us to discard layer activations once updated. This means there is no storage and transportation to an external memory, which is responsible for a large proportion of the power consumption on embedded devices. It is clear the memory advantage increases when fine-tuning more layers. In our setup, the best performance was achieved when fine-tuning the last three layers. This resulted in saving over 81KB of transport to an external memory device. However, the use of the static SGMs introduce computational overhead which can be seen as a trade-off for the memory advantage. Table \ref{tab:memory_stats} shows that the computational overhead for fine-tuning the last three layers using static SGMs is around 80K MACs, which, compared to the total MAC operations for these layers during the forward pass of around 10.3M MACs, is only an increase of less than 1\%. This is mainly due to the fact the SGMs operate on the layer activations, which can be significantly down scaled compared to the layer inputs. For the earlier layers of the model, where the activations are roughly the same size of the inputs, the computational overhead of the SGMs can have a greater impact. Nevertheless, this can be avoided by providing SGMs to groups of layers.

When fine-tuning larger networks with more challenging tasks, the memory advantage will be higher, as these networks have both more layers and larger activations. Also, since the SGMs are static, it is possible to quantize these networks, lowering both the computational and memory overhead. However, this can be hard without an accurate calibration set, which can differ greatly when fine-tuning on tasks which show little resemblance to the pretrained tasks. We will investigate this in future work.

\begin{table}[tb]
    \centering
    \caption{Resource usage of model layers and SGMs.}
    \begin{tabular}{c|ccc|cc}
            & \multicolumn{3}{c|}{Model} & \multicolumn{2}{c}{SGM} \\
        Layer & MAC & Param & Activations & MAC & Param \\
        \hline
        1 & 153,600,000 & 624 & 1,968,128 & 340,480,000 & 1400 \\
        2 & 87,680,000 & 1620 & 430,592 & 74,880,000 & 1400 \\ 
        3 & 35,130,240 & 3250 & 173,056 & 55,206,400 & 5100 \\
        4 & 10,318,080 & 6450 & 77,824 & 39,936 & 624 \\
        5 & 19,456 & 304 & 3072 & 39,936 & 624 \\
        6 & 3,264 & 51 & 192 & x & x
    \end{tabular}
    \label{tab:memory_stats}
\end{table}
\section{Conclusion}
In this work towards memory-efficient on-edge training, we have been able to successfully fine-tune a model using static SGMs. By first training the SGMs on a set of tasks using MAML, while remaining static, these were able to provide accurate updates to fine-tune the model to new similar tasks. By testing our method on a multi-task robot grasping scenario, we showed comparable results to standard backpropagation both for a classification and a regression task. In further work, we will investigate the performance of our method on more challenging tasks like object detection and segmentation. Also, to lower the computational and memory overhead further, we will investigate quantizing the SGMs and the effect on the accuracy of the generated gradients.

\subsubsection{Acknowledgement}
This research received funding from the Flemish Government (AI Research Program).



\clearpage
%
%
\bibliographystyle{splncs04}
\bibliography{egbib}

\begin{thebibliography}{10}
\providecommand{\url}[1]{\texttt{#1}}
\providecommand{\urlprefix}{URL }
\providecommand{\doi}[1]{https://doi.org/#1}

\bibitem{Crafton2019DirectFA}
Crafton, B., Parihar, A., Gebhardt, E., Raychowdhury, A.: Direct feedback
  alignment with sparse connections for local learning. Frontiers in
  Neuroscience  \textbf{13} (2019)

\bibitem{DECONINCK2020103474}
{De Coninck}, E., Verbelen, T., {Van Molle}, P., Simoens, P., Dhoedt, B.:
  Learning robots to grasp by demonstration. Robotics and Autonomous Systems
  \textbf{127},  103474 (2020)

\bibitem{DBLP:journals/corr/FinnAL17}
Finn, C., Abbeel, P., Levine, S.: Model-agnostic meta-learning for fast
  adaptation of deep networks. In: Precup, D., Teh, Y.W. (eds.) Proceedings of
  Machine Learning Research. vol.~70, pp. 1126--1135. PMLR, International
  Convention Centre, Sydney, Australia (06--11 Aug 2017)

\bibitem{frenkel2019learning}
Frenkel, C., Lefebvre, M., Bol, D.: Learning without feedback: Direct random
  target projection as a feedback-alignment algorithm with layerwise
  feedforward training (2019)

\bibitem{han2020extension}
Han, D., Park, G., Ryu, J., jun Yoo, H.: Extension of direct feedback alignment
  to convolutional and recurrent neural network for bio-plausible deep learning
  (2020)

\bibitem{Han2019DirectFA}
Han, D., Yoo, H.J.: Direct feedback alignment based convolutional neural
  network training for low-power online learning processor. 2019 IEEE/CVF
  International Conference on Computer Vision Workshop (ICCVW) pp. 2445--2452
  (2019)

\bibitem{angles}
Hara, K., Vemulapalli, R., Chellappa, R.: Designing deep convolutional neural
  networks for continuous object orientation estimation. CoRR
  \textbf{abs/1702.01499} (2017)

\bibitem{Illing2019BiologicallyPD}
Illing, B., Gerstner, W., Brea, J.: Biologically plausible deep learning - but
  how far can we go with shallow networks? Neural networks : the official
  journal of the International Neural Network Society  \textbf{118},  90--101
  (2019)

\bibitem{DBLP:journals/corr/JaderbergCOVGK16}
Jaderberg, M., Czarnecki, W.M., Osindero, S., Vinyals, O., Graves, A., Silver,
  D., Kavukcuoglu, K.: Decoupled neural interfaces using synthetic gradients.
  In: ICML'17. p. 1627–1635. JMLR.org (2017)

\bibitem{adam}
Kingma, D.P., Ba, J.: Adam: {A} method for stochastic optimization. In: Bengio,
  Y., LeCun, Y. (eds.) 3rd International Conference on Learning
  Representations, {ICLR} 2015, San Diego, CA, USA, May 7-9, 2015, Conference
  Track Proceedings (2015)

\bibitem{launay2020direct}
Launay, J., Poli, I., Boniface, F., Krzakala, F.: Direct feedback alignment
  scales to modern deep learning tasks and architectures (2020)

\bibitem{Lechner2020Learning}
Lechner, M.: Learning representations for binary-classification without
  backpropagation. In: International Conference on Learning Representations
  (2020)

\bibitem{lillicrap2014random}
Lillicrap, T.P., Cownden, D., Tweed, D.B., Akerman, C.J.: Random synaptic
  feedback weights support error backpropagation for deep learning. Nature
  Communications  \textbf{7} (2016)

\bibitem{Manchev2020TargetPI}
Manchev, N.P., Spratling, M.W.: Target propagation in recurrent neural
  networks. J. Mach. Learn. Res.  \textbf{21},  7:1--7:33 (2020)

\bibitem{alex2020theoretical}
Meulemans, A., Carzaniga, F.S., Suykens, J.A.K., Sacramento, J., Grewe, B.F.: A
  theoretical framework for target propagation (2020)

\bibitem{gent_robot}
Molle, P.V., Verbelen, T., Coninck, E.D., Boom, C.D., Simoens, P., Dhoedt, B.:
  Learning to grasp from a single demonstration. CoRR  \textbf{abs/1806.03486}
  (2018)

\bibitem{Moskovitz2018FeedbackAI}
Moskovitz, T.H., Litwin-Kumar, A., Abbott, L.F.: Feedback alignment in deep
  convolutional networks. ArXiv  \textbf{abs/1812.06488} (2018)

\bibitem{nkl2016direct}
N\o{}kland, A.: Direct feedback alignment provides learning in deep neural
  networks. In: Proceedings of the 30th International Conference on Neural
  Information Processing Systems. p. 1045–1053. NIPS'16, Curran Associates
  Inc., Red Hook, NY, USA (2016)

\bibitem{NEURIPS2019_9015}
Paszke, A., Gross, S., Massa, F., Lerer, A., Bradbury, J., Chanan, G., Killeen,
  T., Lin, Z., Gimelshein, N., Antiga, L., Desmaison, A., Kopf, A., Yang, E.,
  DeVito, Z., Raison, M., Tejani, A., Chilamkurthy, S., Steiner, B., Fang, L.,
  Bai, J., Chintala, S.: Pytorch: An imperative style, high-performance deep
  learning library. In: Wallach, H., Larochelle, H., Beygelzimer, A.,
  d\textquotesingle Alch\'{e}-Buc, F., Fox, E., Garnett, R. (eds.) Advances in
  Neural Information Processing Systems 32, pp. 8024--8035. Curran Associates,
  Inc. (2019)

\bibitem{DBLP:journals/corr/LongSD14}
Shelhamer, E., Long, J., Darrell, T.: Fully convolutional networks for semantic
  segmentation. IEEE Trans. Pattern Anal. Mach. Intell.  \textbf{39}(4),
  640–651 (Apr 2017)

\bibitem{Verhelst2020}
Verhelst, M., Murmann, B.: Machine Learning at the Edge, pp. 293--322. Springer
  International Publishing, Cham (2020)

\end{thebibliography}
\end{document}